\documentclass[sn-mathphys,Numbered]{sn-jnl}

\usepackage{graphicx}%
\usepackage{multirow}%
\usepackage{amsmath,amssymb,amsfonts}%
\usepackage{amsthm}%
\usepackage{mathrsfs}%
\usepackage[title]{appendix}%
\usepackage{xcolor}%
\usepackage{textcomp}%
\usepackage{manyfoot}%
\usepackage{booktabs}%
\usepackage{algorithm}%
\usepackage{algorithmicx}%
\usepackage{algpseudocode}%
\usepackage{listings}%

\usepackage{graphics}

\begin{document}

\title[Article Title]{Multi-Resolution Haar Network:\par Enhancing human motion prediction via Haar transform}


\author[1]{\fnm{Li} \sur{Lin}}




\affil[1]{\orgdiv{School of Computer Engineering and Science}, \orgname{Shanghai University}, \orgaddress{\street{Shangda Road}, \city{Baoshan}, \postcode{200444}, \state{Shanghai}, \country{China}}}


\abstract{The 3D human pose is vital for modern computer vision and computer graphics, and its prediction has drawn attention in recent years. 3D human pose prediction aims at forecasting a human's future motion from the previous sequence. Ignoring that the arbitrariness of human motion sequences has a firm origin in transition in both temporal and spatial axes limits the performance of state-of-the-art methods, leading them to struggle with making precise predictions on complex cases, e.g., arbitrarily posing or greeting. To alleviate this problem, a network called HaarMoDic is proposed in this paper, which utilizes the 2D Haar transform to project joints to higher resolution coordinates where the network can access spatial and temporal information simultaneously. An ablation study proves that the significant contributing module within the HaarModic Network is the Multi-Resolution Haar (MR-Haar) block. Instead of mining in one of two axes or extracting separately, the MR-Haar block projects whole motion sequences to a mixed-up coordinate in higher resolution with 2D Haar Transform, allowing the network to give scope to information from both axes in different resolutions. With the MR-Haar block, the HaarMoDic network can make predictions referring to a broader range of information. Experimental results demonstrate that HaarMoDic surpasses state-of-the-art methods in every testing interval on the Human3.6M dataset in the Mean Per Joint Position Error (MPJPE) metric.
All codes are available at \url{https://github.com/xhaughearl/HaarMoDic}.
}




\maketitle

\section{Introduction}

Human motion prediction is typically formulated as a sequence transduction problem, which takes in the historical pose sequence and derives future motion. It is vital for several applications, including building robust augmented reality(AR) system\cite{makris2016augmented}, supporting athletes \cite{honda2020rnn}, and autonomous driving \cite{Gulzar2021Survey,sun2022p4p}.

\begin{figure}[t]
    \centering
    \includegraphics[width=0.9\textwidth]
    {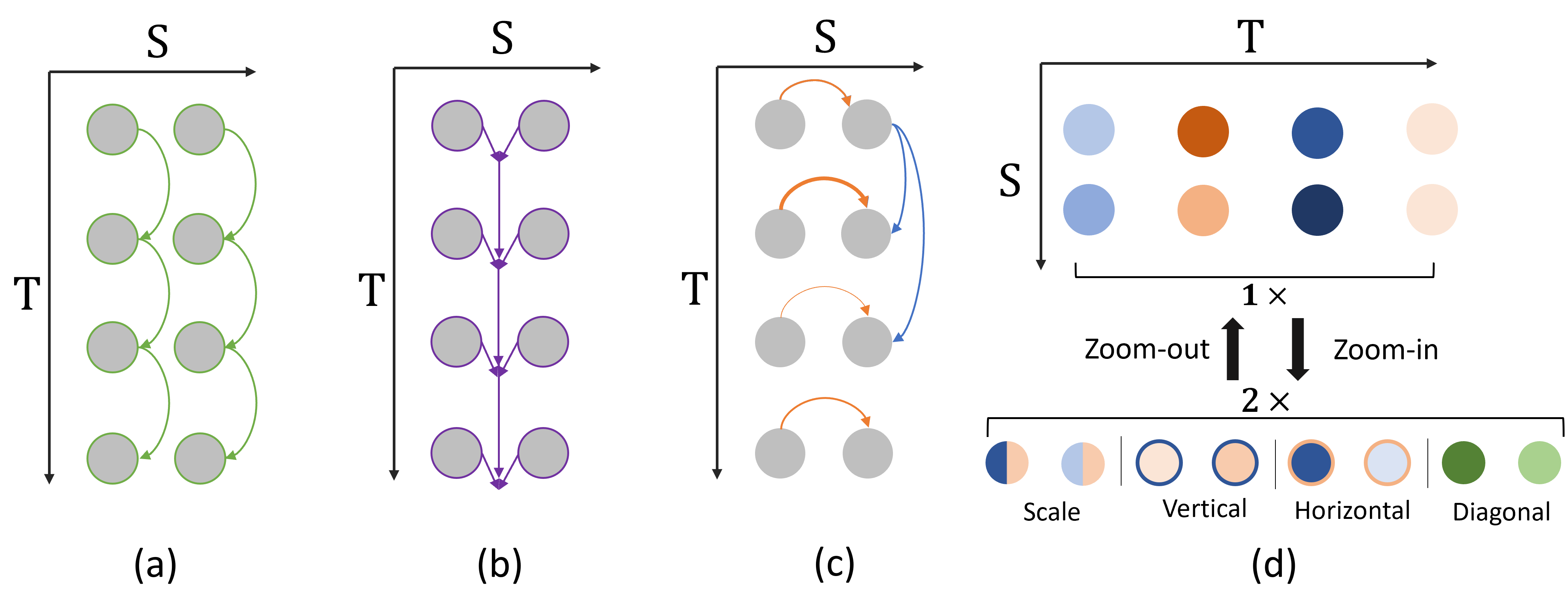}
    \caption{\textbf{Visualization of different methods on the virtual graph of joints.} Every circle above denotes a single joint which allocates by frame \textbf{T} and position \textbf{S}. \textbf{T} on the axes denotes the temporal axis, representing the transition frame by frame, and \textbf{S} denotes the spatial axis, on behalf of the structural representation of the human body. Approaches vary in how to leverage temporal and spatial information. \textbf{(a)} RNN-based methods focus on the transition within the temporal axis. \textbf{(b)} GCN-based methods first weigh different elements at the same spatial layer and then regress them through the temporal axis. \textbf{(c)} Attention-based methods use attention mechanisms to represent the whole virtual graph, as shown in blue and brown arrows, representing temporal and spatial attention. \textbf{(d)} The proposed methods in this paper use the 2-dimensional Haar transform to transfer the whole graph into a new and high-resolution coordinate.}
    \label{fig:vir_img}
\end{figure}

Current approaches have made significant progress in addressing the challenge of human motion prediction. However, a common issue is the difficulty in balancing the spatial and temporal information, which limits the effectiveness of state-of-the-art methods. Human pose sequences can be visualized as virtual graphs with two dimensions, one represent for the relative spatial structure of joints, the other is for the temporal propagation through timeline, as shown in Fig \ref{fig:vir_img}. Recurrent neural network (RNN) \cite{martinez_human_2017,fragkiadaki_recurrent_2015,cui2020learning} has been commonly used for human motion prediction due to the nature of sequence forecasting, with modules such as Long Short-Term Memory (LSTM)\cite{hochreiter1997long} and Gated Recurrent Unit (GRU)\cite{chung2014empirical} being introduced. However, they struggle to capture the structural information of human skeletons and are unreliable in predicting distant future motions \cite{guo_back_nodate}. GCN-based feed-forward architectures  \cite{mao_learning_2019,dang_msr-gcn_2021} have been proposed to address spatial neglecting issues, but they typically require extra knowledge about the human structure or have difficulty deriving a correct skeleton from generic graphs. Recently, attention mechanisms have been utilized to identify temporal similarity and compose predictions from past transitions or to combine "spatial/temporal attention" to constrain both axes \cite{mao2020history,aksan2021spatio}. However, these methods fail to surpass the performance of GCN-based methods in long-term prediction due to the intentional separation of the spatial and temporal parts, resulting in a trade-off between spatial and temporal information and a trade-off between short-term and long-term prediction.

To alleviate the above limitations, HaarMoDic is proposed in this paper, a novel network that utilizes the Haar transform to enhance the extraction of information. The Haar transform simultaneously enhances the quality of mining information in spatial and temporal axes without trading off between long-term and short-term prediction.

The Haar Transform associates finite adjacent elements in order and is part of the wavelet transform family. Instead of mining information within the given spatial and temporal axes, the whole graph can be projected to a mixed-up coordinate in higher resolution to gain detailed information hidden in the two directions. A virtual graph can be decomposed into four components by four wavelet functions, as shown in Fig \ref{fig:vir_img}(d). Two kinds of scenarios can be defined, namely, zoom-in and zoom-out scenarios. By applying the 2D Haar transform, a zoom-in scenario can be conducted on the virtual graph. The coordinate with two times resolution without losing information is where the Haar transform aiming for. To achieve a zoom-out scenario, it only needs to operate an inverse 2D Haar Transform on the virtual graph to be zoomed out. These processes can be conducted recursively, and each time a zoom-in scenario is conducted, the adjacent area in the scope becomes wider. The Haar transform satisfies the two requirements above: (i) the constraint of two axes simultaneously and (ii) mining in a mixed-up manner. Hence, it allows the network to give a better scope to temporal and spatial information based on the Haar coefficients. To the best of our knowledge, it is the first approach to introduce the Haar transform to tackle the task of human motion prediction.

 HaarMoDic stands for using \textbf{Haar} Transform to enhance \textbf{Mo}tion pre\textbf{Dic}tion. The HaarMoDic network benefits from a series of transforms, including the Haar transform and Discrete Cosine Transform (DCT). There are several DCT modules to project coordinate between the original XYZ coordinate and a smooth trajectory space back and forth at the top and bottom of the HaarMoDic network. The full-connected layers that conduct the pre-/post-processing also help to generate the finest performance. The most complexity of HaarMoDic is kept inside the Multi-Resolution Haar blocks (MR-Haar block). In an MR-Haar block, we maintain several roads of pipeline in different resolutions. By conducting the zoom-in and zoom-out scenarios, the spectrum fed in shift from different roads of pipelines. Each road of pipeline equips with a sequence of full-connected (FC) blocks comprised of several full-connected layers and layer-normalization operations. All roads of feature maps are merged at the bottom of the MR-Haar block, which is visualized as a trapezoidal shape.


To summarize the technical contribution of this paper,
\begin{quote}
 \begin{itemize}
 \item 	It proposes a method to elaborate on mining the information from temporal and spatial axes, applying the Haar transform to extract information explicitly to avoid the trade-off between \textbf{Spatial} and \textbf{Temporal} axes.
 
\item 	It proposes the HaarMoDic network for human motion prediction. HaarMoDic consists of Haar blocks combined with Haar transform ladders and MLP-based regressors. HaarMoDic can effectively learn spatial and temporal representation in a multi-resolution manner.

\item HaarMoDic achieves state-of-the-art performance on the benchmark of Human3.6M \cite{ionescu2013human3} , benefits from the multi-resolution structure.   
 \end{itemize}
\end{quote}

\section{Related work}

Human motion prediction is formulated as a sequence transduction problem, inputting previously historical motion to anticipate future motion transition. 

The introduction of RNN \cite{fragkiadaki2015recurrent,jain2016structural,liu2019towards,martinez2017human,chiu_action-agnostic_2019,honda2020rnn} and GCN \cite{dang_msr-gcn_2021,guomulti,li2020dynamic,mao2020history,mao2019learning,ma2022progressively,li2021symbiotic} modules generate more acceptable results than conventional methods \cite{lehrmann2014efficient,wang2005gaussian,taylor2007modeling}, with some methods \cite{cai2020learning,mao2019learning,aksan2021spatio} using the transformer to tackle tasks and produce acceptable output with attention mechanisms. In recent years, there has been a sophisticated method \cite{guo_back_nodate} which utilizes the MLP to gain cutting-edge performance.

\textbf{RNN-based methods.}
In the deep learning era, the Recurrent neural network (RNN) was introduced to the domain of human pose prediction, including the Long-short time memory (LSTM) and Gated recurrent unit (GRU) blocks. By variating the arrangement of each RNN module, a handful of methods generate confidential results on human motion prediction.
Fragkiadaki et al. \cite{fragkiadaki_recurrent_2015} consolidate the effectiveness of the encoder-recurrent-decoder framework, which utilizes nonlinear encoder and decoder networks and LSTM to update latent space. However, they are only train action-specific-model, i.e., one model for one pose class.

Using residual architecture, Martinez et al. \cite{martinez_human_2017} propose a model that contains a sequence of GRUs and is capable of propagating through the recurrent chain. Residual architecture prolongs the receipt field of the sequence of the model input, i.e., more frames are now in the scope when training and testing the model. Besides, it is a multi-action model, i.e., a model trained on different types of poses, digging out the regularities across poses with larger datasets. Furthermore, it supervises the joints' velocity, which generates smoother results to increase the performance. 

However, the chain of GRU or LSTM in the approaches above accumulates errors, and the longer the sequence, the more errors are inherited \cite{dang_msr-gcn_2021}. Besides, it suffers from poor parallelity and consumes much memory \cite{guo_back_nodate}. To fix the issue, Chiu et al.  \cite{chiu_action-agnostic_2019} propose a concurrency architecture, utilizing three roads with different densities of LSTM to capture the temporal information in the input sequence. \cite{butepage2017deep,butepage2018anticipating} use sliding windows, \cite{hernandeziccv2019,li2018convolutional} use convolutional operation, and \cite{gui2018adversarial} utilizes adversarial training to tackle with the issue. Nevertheless, these fixes cannot patch all the shortcomings of the RNN chain structure. A basic reason is that the RNN structure is not explicitly designed to extract the spatial information of human motion sequence, which trade off the temporal information with spatial ones.

\textbf{GCN-based methods.}
Recent work utilizes the Graph Convolution Network (GCN) to dig out spatial connections of joints and the hidden feature in the trajectory space of human pose. It is intuitive to relate human motion with GCN. Features of points and graphs originating in human motion's nature make GCN a go-to choice to extract spatial information.

Mao et al. \cite{mao_learning_2019} propose a feed-forward network instead of RNN, which avoids the accumulation of errors. This work first introduces the GCN to human pose prediction. Mao et al. \cite{mao_history_2020} further improve the previous work by enhancing the temporal encoding by slicing the historical human motion sequence into sub-sequences and utilizing the attention mechanism to capture the similarity between the current context and historical sub-sequence. A GCN-based predictor is applied to regress the human pose at the final stage.

Dang et al. \cite{dang_msr-gcn_2021} propose an end-to-end model called MSR-GCN, which replace the part of the DCT transform by extracting feature from coarse to fine. \cite{ma_progressively_2022} propose a two-stage architecture that views initial guess as a support to gain a more accurate prediction. 

Even though the modules with GCN generate qualified results, they still stumble in complex cases and leave room for performance. The balance of temporal and spatial information should be well-tuned to find the sweat point between them.

\textbf{Attention-based methods.}
Transformers gradually become a go-to choice for constructing a network structure. Some works introduce the attention mechanism to deal with the task. Mao et al. \cite{mao2020history} use attention to find the temporal similarity in a long range. Aksan et al. \cite{aksan2021spatio} takes not only temporal consistency but also the pairwise relationship of joints by combining "spatial attention" and "temporal attention."

Nevertheless, the trade-off between spatial and temporal information limits their performance. In the benchmark of Human3.6M, \cite{mao2020history} cannot even surpass the previous works in long-term prediction.

Besides the attention-based methods, simple designs based on multi-layer perception (MLP) made considerable progress. Guo et al. \cite{guo_back_nodate} propose an MLP-based network to deal with human motion prediction. The structure of their work is straightforward compared to former works. However, because of the static dimension of every layer of inner MLP, this method also suffers from the trade-off problem.

\section{Our approach}
The overview of the architecture of HaarMoDic will be introduced at section \ref{overview_archi}. To fully discuss the Haar transform applied in this paper, we discuss a basic form and a 2-dimensional Haar transform at section \ref{Haar_transform} and section \ref{2D_Haar_transform}. Then the significent contributing module, Multi-Resolution Haar Block will be introduced at section \ref{MR-Haar_block}. Finally, the loss witch implemented will be discussed at section \ref{LOss}.
\begin{figure}[t]
    \centering
    \includegraphics[width=0.99 \textwidth]
    {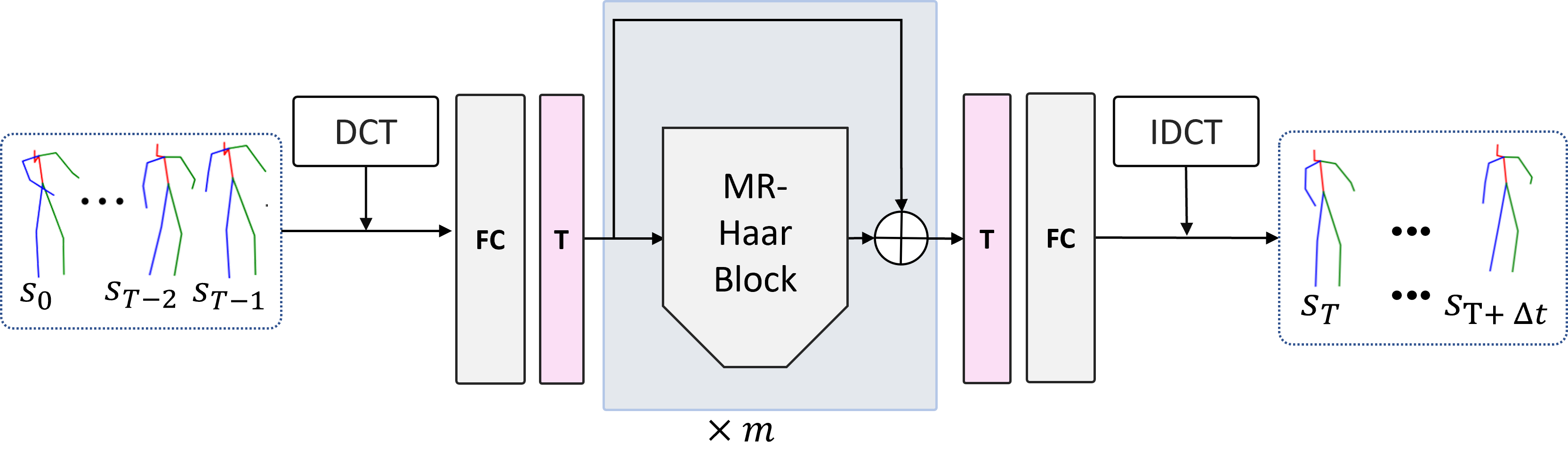}
    \caption{\textbf{An overview of the architecture of HaarMoDic.}  \textbf{FC} denotes for full-connected layer, and \textbf{T} denotes for transpose operation. \textbf{DCT} and \textbf{IDCT} denote discrete cosine transform and the inverse operation. The detailed explanation of the \textbf{MR-Haar Block} is in section \ref{MR-Haar_block}. To get better performances, $m$ MR-Haar blocks are concatenated in the network.
    }
    \label{fig:overview_of_TAT}
\end{figure}
\subsection{An overview of HaarMoDic}
\label{overview_archi}

The proposed network consists of three main parts: 
Multi-Resolution Haar (MR-Haar) block, DCT/IDCT module, and pre/post process block. We will discuss the MR-Haar block in detail in section \ref{MR-Haar_block}. The pre and post-process blocks combine full-connected layers and transpose operations. As Fig \ref{fig:overview_of_TAT}, it first projects the joint coordinates to the spectrum by DCT transform. Then, it pre-processes by a full-connected layer and transposes the given DCT spectrum. Several MR-Haar blocks are followed to extract the information from the spatial and temporal axes, which are the core of the HaarMoDic. Each MR-Haar block passes residuals to the next block, which avoids gradient vanishing. Finally, it post-processed the last return of the MR-Haar block by transposing the output spectrum and feeding it into a full-connected layer. Applying the inverse discrete cosine transform reaches the predicted joint coordinates. 
\begin{figure}[t]
    \centering
    \includegraphics[width=0.89 \textwidth]
    {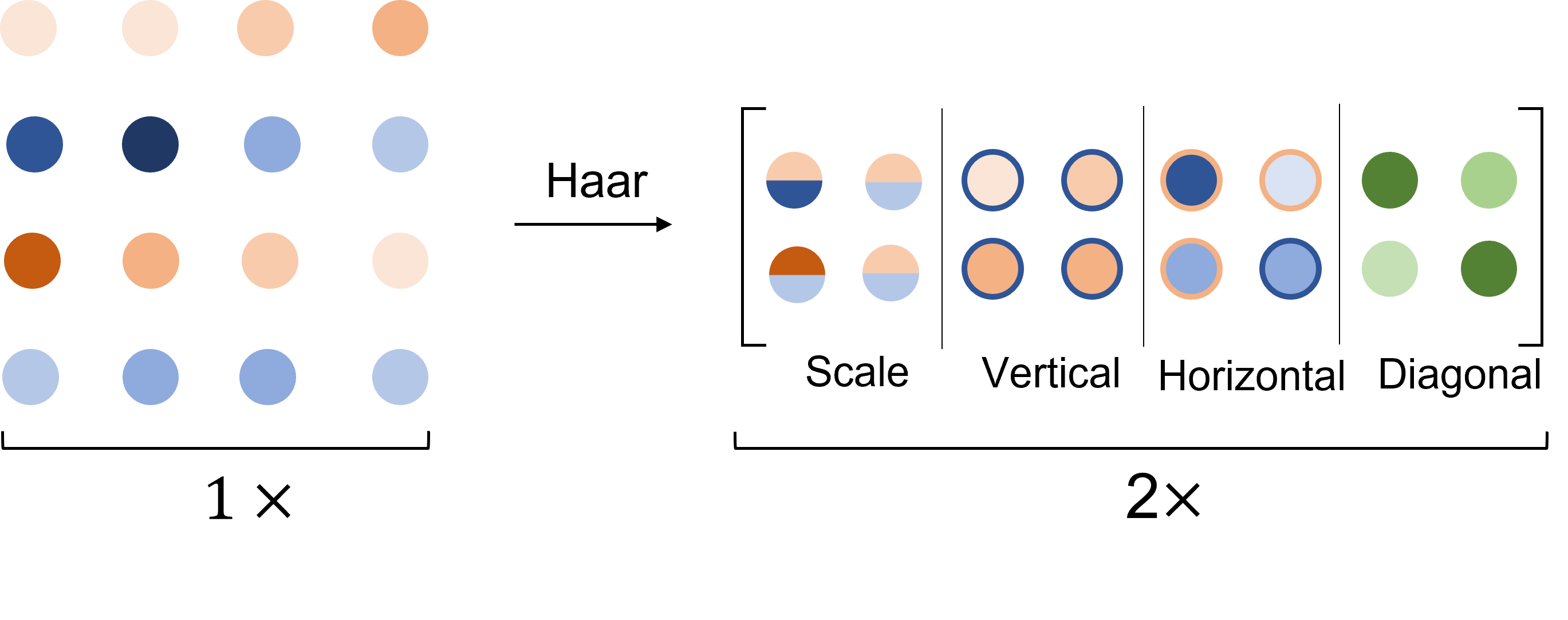}
    \caption{\textbf{Visualization of applying 2D Haar transform over the virtual graph.} With the nature of the 2D Haar transform, the virtual graph comprises four matrices: Scale, Vertical, Horizontal, and Diagonal coefficients matrices. Scale matrix consists of the average sums, represented as circles with two colors half-half. Vertical matrix mainly contains the vertical high-frequency information, depicted as inner brown and outer blue. Similarly, the Horizontal matrix shows as inner blue and outer brown. The diagonal matrix contains a mixture of diagonal information, represented in green. Hence, we can project the motion sequence coordinate to a two-times high-resolution coordinate.}
    \label{fig:haar_on_virtual_graph}
\end{figure}
\subsection{Haar Transform}
\label{Haar_transform}
The Haar transform is one of the wavelet transforms \cite{cotter_2020}, which decomposes the original signal and conducts multi-scale analysis. To achieve these, the scale function and mother function are needed. The Haar scaling function is a down-sampling function that maintains the low-frequency part from the original signal. On interval $[\gamma_1,\gamma_2)$, it is defined as
$$
h_1(x) = \left\{ \begin{array}{cl}
1 & : \ x \in [\gamma_1,\gamma_2), \\
0 & : \ \text{elsewhere,}
\end{array} \right.
$$
and the mother function over the interval $[\gamma_1,\gamma_2)$ is 
$$
h_2(x) = \left\{ \begin{array}{cl}
1 & : \ x \in  [\gamma_1,\frac{\gamma_1 + \gamma_2}{2}), \\
-1 & : \ x \in [\frac{\gamma_1 + \gamma_2}{2},\gamma_2) ,\\
0&:\text{elsewhere.}
\end{array} \right.
\\
$$

In a discrete situation, supposing the signal with length $n$, which $f\in L(\mathbb{Z}_n)$. The discrete Haar transform is as
$$
H = \left\{ \begin{array}{cl}
f(2j+1) + f(2j) & : \ 0 \le j\le \frac{n}{2}-1 \\
f(2j+1) - f(2j) & : \  \frac{n}{2} \le j\le n-1 .
\end{array} \right.
$$
The two parts in combination $H$ represent the approximating and scaling factor. They are filtered by scaling and mother function.

As for the human motion prediction, the reason why Haar transform is vital for improving the performance is that this family of transforms can effectively change the resolution of the original data while preserving all the information without losing details. This feature is critical for a mission with little tolerance of error accumulation like prediction of human motion.
\subsection{2-Dimensional Haar transform}
\label{2D_Haar_transform}

2-dimensional Haar transform is utilized to mix the adjacent parameters of joints and arrange them in a high-resolution order as fig \ref{fig:haar_on_virtual_graph}. The network adjusts this process to change the spectrum's resolution, yielding cutting-edge performance. The 2D Haar transform is a lossless, size-changing process, as a expansion of 1D version of that. Stacking all the sub-matrices in the same axes increases the resolution of the DCT spectrum. Splitting a matrix into four parts and conducting inverse haar transform gives it a matrix in lower resolution.

A 2-dimensional discrete Haar transform can be defined in a recursive matrix convolution manner. Suppose there is a matrix $g^n$ and convolution kernel $\omega$, where $g^n \in \mathbb{R}^{H \times W}$ and $\omega \in \mathbb{R}^{2 \times 2}$. It denotes a convolution with stride as $\ast$, which
$$
\hat{g}^{n+1}[x,y] = (g^n \ast \omega)[x, y] = \sum_{i} \sum_{j} g^n[2x + i,2y + j] \cdot \omega[i, j],
$$
where $\hat{g}^{n+1} \in \mathbb{R}^{\frac{H}{2} \times \frac{W}{2}}$.

There are four kinds of kernel in 2D discrete Haar transform, which are defined as
$$
\begin{matrix}
 \omega_s=\begin{pmatrix}\quad1 &\quad1  \\\quad1 &\quad1 \end{pmatrix} &&
 \omega_h=\begin{pmatrix}\quad1 &-1  \\\quad1 &-1 \end{pmatrix} &&
 \omega_v=\begin{pmatrix}\quad1 &\quad1  \\-1 &-1 \end{pmatrix} &&
 \omega_d=\begin{pmatrix}\quad1 &-1  \\-1 &\quad1 \end{pmatrix}
\end{matrix},
$$
and each is represented by scaling, horizontal, vertical, and diagonal kernel. Every time the 2D Haar transform is applied, it will get one matrix for low-frequency information denoted as $g_s$ and three coefficients matrix containing high-frequency information denoted as $g_h,g_v,g_d$ as
$$
\begin{matrix}
\hat{g}^{n+1}_s=\frac{1}{2} g^{n} \ast \omega_s &&
\hat{g}^{n+1}_h=\frac{1}{2} g^{n} \ast \omega_h &&
\hat{g}^{n+1}_v=\frac{1}{2} g^{n} \ast \omega_v &&
\hat{g}^{n+1}_d=\frac{1}{2} g^{n} \ast \omega_d
\end{matrix},
$$
and it concatenates four sub-matrices to derive for $g^{n+1}$ as 
$$
g^{n+1}=[\hat{g}^{n+1}_s|\hat{g}^{n+1}_h|\hat{g}^{n+1}_v|\hat{g}^{n+1}_d].
$$

For the Inverse 2D Haar transform, it defines four kernels as a set of $Kernels$. To inverse the operation, the slice of the original matrix is derived as 
$$
g^{n}[i:i+1,j:j+1]=\sum_{k}^{Kernels} \hat{g}^{n+1}_k[\frac{i}{2},\frac{j}{2}]\cdot \omega^T_k ,
$$
where $k$ is each convolution kernel same as 2D Haar transform.

Under the background of boosting motion prediction, each kernel was attributed its own function during processing input spectrum. To access a macro scope of information, $\omega_s$ functions as an equivalent to a linear filter, which allow more information bring into the scope. To extract the inner relationship amid the temporal and spatial axes, $\omega_h$ and $\omega_v$ are utilized to generate the differential through two axes and present the inner information to the FC blocks latter. Finally, $\omega_d$ is designed for preserve the partial information which neglect by the convolution with $\omega_h$ and $\omega_v$. 

\subsection{MR-Haar Block}\label{MR-Haar_block}

\begin{figure}[t]
    \centering
    \includegraphics[width=0.85 \textwidth]
    {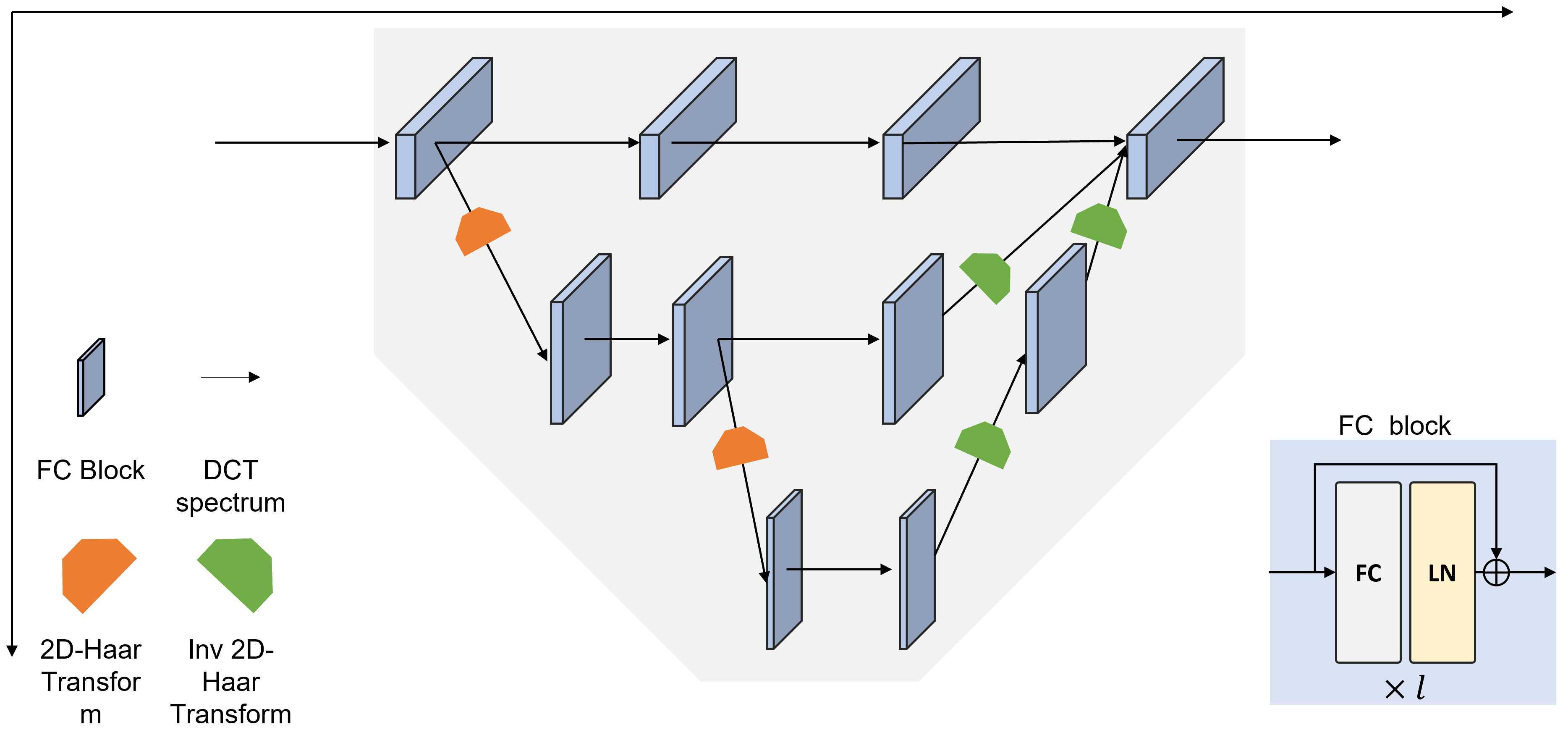}
    \caption{\textbf{Illustration of MR-Haar Block.} This figure illustrates one kind of MR-Haar block containing three orders of Haar transform, and the number of the full-connected layers for each order are 4, 2, 1. \textbf{FC} denotes the full-connected layer, and \textbf{LN} denotes for layer normalization. All the spectrums in different orders are merged together at the second from the last stage. The residual pathway is not included.
    }
    \label{fig:mr_haar_block}
\end{figure}

As shown in Fig \ref{fig:mr_haar_block}, the MR-Haar comprises two major elements: the 2-dimensional Haar transform and the Full-connected (FC) blocks.

The key to completely utilizing the aids of the scope of 2D Haar transform is to maintain different resolutions through the pipelines. Inspired by HR-Net \cite{wang2020deep}, the module's shape contains several branches, each counting for different resolutions. It starts at a low-resolution branch and gradually expands the number of high-resolution branches. Finally, all the branches will be merged back to the original resolution.
Any input feature map faces two scenarios: \textbf{go through FC blocks} or \textbf{zoom in/out}. 

The scenario of keeping the same resolution, neither zoom-in nor zoom-out, is just through a sequence of full-connected layers. Suppose the original input spectrum denotes as $G_0^0$. The lower title is the index of how many operations are applied. The upper title shows the level of resolution; the higher the number, the higher the resolution. As far as in the same resolution, no Haar transfer is applied; it passes through a sequence of fully connected blocks ($FC$) without changing the shape like
$$G_0^0  \to   G_1^0 \to G_2^0\Longleftrightarrow G_0^0  \xrightarrow{\text{FC}}   G_1^0 \xrightarrow{\text{FC}} G_2^0,$$
where the Layer Normalization inside the FC block is defined as  
$$
L_{layer}(x)=\frac{x_i-\mu}{\sqrt{\sigma^2+\epsilon}}\times\gamma+\beta,
$$
where $\gamma$ and $\beta$ are the learnable parameters, while $\mu$ and $\sigma$ denotes for mean and variation of the $x$ sequence.
Given the structure of FC block, a single transition is as
$$
G_1^0 = L_{layer}(w^T\cdot G_0^0 + b^T),
$$
where $w$ and $b$ are the inner parameters of the full-connected layers.

In zoom-in scenarios, suppose that it starts at $G_0^0$, which is like
$$
\begin{matrix}
\begin{matrix}
G_0^0 & \   \\
 & \searrow   \\
 &  & G_1^1& 
\end{matrix} \Longleftrightarrow   G_0^0 \xrightarrow{\text{Haar}} G_1^1
\end{matrix}.
$$
In this scenario, supposing the original space is $G_0^0 \in \mathcal{R}^{S\times T}$, it is 
projected to the new coordinate which is $G_0^0 \in \mathcal{R}^{2S\times \frac{T}2}$. We choose to stack the extra resolution on the S axis. It should be noticed that the network do not conduct zoom-in scenario on the spatial axis but choose the axis, which was the spatial axis, to hold the place for the extra expanding resolution. The new coordinate does not completely inherit the original spatial meaning, and the scope mixes both spatial and temporal manner.
The Zoom-in scenario is defined with 2D Haar transform as 
$$
G_0^1 = \frac12[G_0^0\ast \omega_s|G_0^0\ast \omega_v|G_0^0\ast \omega_h|G_0^0\ast \omega_d]
$$
Similarly, an inverse operation of 2D Haar transform is required when zooming out the feature map. We slice the input feature map into four blocks with the same size which it was compounded; then, it derive the low-resolution feature map as
$$
\begin{matrix}
\begin{matrix}
   &      &&G_2^0 & \\
 && \nearrow   &&  \\
&G_1^1 &
\end{matrix} \Longleftrightarrow   G_1^1 \xrightarrow{\text{Inv Haar}} G_2^0.
\end{matrix}
$$
It is intuitive to manually divide the whole DCT spectrum block at resolution level $n$ into four coefficient blocks as 
$$
G^n = [g_s|g_v|g_h|g_d].
$$
Each of the sub-matrices are applied an inverse Haar transform to derive the zoom-out spectrum,
$$
G^{n-1}[i:i+1,j:j+1]=\sum_{k}^{Kernels} g_k[\frac{i}{2},\frac{j}{2}]\cdot K^T_k ,
$$

Any variation of the Multi-resolution Haar block should maintain the same shape as the feature map from end to end. The sum of zoom-in scenarios should equal the zoom-out scenarios. the structure should be similar to
$$
\begin{matrix}
G_0^0 & \to  & G_1^0 &\to& G_2^0 \\
 & \searrow  && \nearrow  \\
 &  & G_1^1
\end{matrix}.   
$$

\subsection{Loss}
\label{LOss}
Following the manner of implementing loss as \cite{guo_back_nodate}, instead of the coordinate of joints from scratch, the network supervises the residual of joints from the last frame, which is as
$$
\mathscr{F}(s_0,s_1,\cdots,s_{T-1}) \to (s'_{T},\Delta s'_{T+1},\cdots,\Delta s'_{T+\Delta t - 1}).
$$
It derives the final result by adding the residual back to the last frame,
$$
s'_{t} = s'_T+\Delta s_t,\quad t\in(T,T+\Delta t)
$$
\textbf{Objective function.} Our objective function $\mathcal{L}$ contains two main components, the regular loss $\mathcal{L}_{re}$ and velocity loss $\mathcal{L}_{v}$, where
$$
\mathcal{L} = \mathcal{L}_{re} + \mathcal{L}_{v}.
$$
$\mathcal{L}_{re}$ is about the distance between the prediction and the grand truth, while $\mathcal{L}_{v}$ is for the velocity.

The regular loss is the relative Euclidean distance between the joints of grand truth and prediction, which is 
$$
\mathcal{L}_{re}=\frac{1}{\Delta t}\sum_{}^{}\left\| s'_{T:T+\Delta t}-s_{T:T+\Delta t} \right\|_2.
$$
When calculating the $\mathcal{L}_{re}$, each skeleton is decomposed to the coordinate form.

The velocity loss is an auxiliary objective function that supervises the discrete velocity between grand truth and prediction. We use discrete subtraction as the discrete velocity as 
$$
v_i = x_{i+1}-x_i.
$$
It can calculate discrete velocity every two frames for each joint as 
$$
V_i=s_{i+1}-s_i=\sum_{n=0}^{N-1}J^i_{n+1}-J^i_n.
$$
Hence, the network can supervise the velocity as 
$$
\mathcal{L}_v=\frac{1}{\Delta t-1}\sum_{}^{}\left\| V'_{T:T+\Delta t -1}-V_{T:T+\Delta t-1} \right\|_2.
$$

\section{Experiments}
Massive experimental results demonstrate that HaarMoDic surpass state-of-the-art methods on Human3.6M dataset. The Human3.6M dataset and the evaluation metric for the dataset will be introduced at section \ref{dataset}. Then the comparison settings will be described in details at section \ref{comp}. The results of quantitative and qualitative experiments will be discuss at section \ref{quantita}, and ablation study is demonstrated at section \ref{ablation}.

\subsection{Dataset and evaluation metric}
\label{dataset}
\subsubsection{Human3.6M dataset}
Human3.6M dataset\cite{ionescu2013human3} comprises 7 individuals engaged in 15 distinct actions, with 32 joints annotated for each pose. The evaluation procedure aligns with the other approaches listed in table.\ref{table:tab_res_h36m}, where subset S5 of the Human3.6M serves as the test set, S11 as the validation set, and the remaining sets as the training data. Prior works have employed varying testing sampling approaches, such as 8 samples per action , 256 samples per action, or utilizing all test set samples. Since 8 samples are insufficient and including all test samples would not ensure a balanced representation across diverse actions and sequence lengths, we opt for 256 samples per action for testing. Our evaluation covers 22 joints, consistent with the other approach in setup comparison.
\subsubsection{Evaluation Metrics}

It comes naturally to abstract human poses as the skeletons, representing several key points of joints. Hence, human motion prediction is a process that predicts future joint positions by observing the past joint coordinates. This paper focuses on the prediction in 3-dimensional space where a 3D vector represents each joint. Each dimension of the vector represents one of the XYZ coordinates.
Given a $L$-length input array $X$, where every three adjacent elements construct a complete XYZ coordinate, is as
$$
X = (x_0,x_1,\cdots,x_L).
$$

A joint is denoted as $J$, where $J \in \mathbb{R}^3$. For a joint that is indexed as $i$, we map the input sequence $X$ with the joint as,
$$
J_i = (x_{3i},x_{3i+1},x_{3i+2}),
$$
where the $x$ denotes for the each axis in XYZ system and $i \in [0,\frac{L}{3}-1]$.

A single skeleton denotes $s$, supposing that a skeleton containing $N$ joints , thus $s \in \mathbb{R}^{N\times 3}$. Giving a skeleton sequence with length $T+\Delta t$, indexing frames from $0$ to $T+\Delta{t} - 1$, a full snippet of sequence formulates as 
$$
s_i = (J_0^i,J_1^i,\cdots,J_{N-1}^i) \quad i\in\left[ 0,T+\Delta{t} - 1 \right].
$$
Defining the first $T$ frames as a historical sequence, the target is to predict the coordinate of the joints in the next $\Delta{t}$ frames, which are formulated as
$$
\mathscr{F}(s_0,s_1,\cdots,s_{T-1}) \to (s_{T},s_{T+1},\cdots,s_{T+\Delta t - 1}).
$$

This paper reports the Mean Per Joint Position Error (MPJPE) as testing metric, a standard metric extensively employed to assess 3D pose accuracy. This measurement computes the average Euclidean distance between the predicted and ground-truth 3D joint coordinates across various joints as 
$$
Mean(\sum_{i=0}^3(x'_i-x_i)^2,\quad x\in s\quad x'\in s')
$$
To evaluate the prediction with different intervals, we measure the 8 frames after the last frame given with a time interval of 80,160,320,400,560,720,880,1000 milliseconds. 

\subsection{Comparison settings}
\label{comp}
In testing scenarios, the input length is established as $T$ = 50, and for the output, predict future $10$ frames. When evaluating our model, an auto-regressive approach is adopted to generate motion for extended timeframes. The feature dimension, denoted as C, is determined as $3\times K$, where $K$ represents the number of joints. Specifically, for the Human3.6M dataset, $K$ is set to 22.

To train our network, we choose 256 as the batch size, and we employ the Adam optimizer. We adopt the Pytorch Wavelet module \cite{cotter_2020} to accelerate the computation of the Haar transform within the Pytorch framework, executed on a single NVIDIA RTX 3090 graphics card.
For training on the Human3.6M dataset, we subject the network to 80,000 iterations. The learning rate commences at 0.0003 and decreases to 0.00006 after 30,000 iterations. After 30,000 iterations, the learning rate times 0.85 for every 3,000 iterations. 
Throughout the training process, we apply data augmentation through front-back flips and sequence reverse. The sequence reverse augmentation technique introduces randomness by flipping motion sequences during training. Front-back flips augmentation randomly flips the input skeleton from front to back.


\begin{table}
    \centering
    \renewcommand\arraystretch{1.2}
    \setlength{\aboverulesep}{0pt}
    \setlength{\belowrulesep}{0pt}
    \caption{\textbf{Results on Human3.6M} for different prediction time steps~(ms). We report the MPJPE error in \textit{mm} for each method. Lower is better. 256 samples are tested for each action. $\dag$ indicates that the results are taken from the paper~\cite{mao2020history,ma2022progressively}. Note that ST-DGCN~\cite{ma2022progressively} uses two different models to evaluate their short-/long- term performance, here we report their results of a single model that performs better in the long term for fair comparison. }
    \label{table:tab_res_h36m}
    \begin{tabular}{l|cccccccc} 
    \toprule
             & \multicolumn{8}{c}{MPJPE (millimeter) $\downarrow$}                                                                                                                                                                                                                                                                                                               \\
    Time (ms)                                                                                           & 80                                       & 160                                      & 320                                      & 400                                      & 560                                       & 720                                       & 880                                       & 1000                                       \\ 
    \midrule
    Res-RNN  (2017) $\dag$ \cite{martinez2017human}          & 25.0 & 46.2 & 77.0 & 88.3 & 106.3 & 119.4 & 130.0 & 136.6  \\
    convSeq2Seq (2018) $\dag$ \cite{li2018convolutional}    & 16.6                                     & 33.3                                     & 61.4                                     & 72.7                                     & 90.7                                      & 104.7                                     & 116.7                                     & 124.2                                      \\
    LTD-50-25 (2019) $\dag$ \cite{mao2019learning}          & 12.2 & 25.4 & 50.7 & 61.5 & 79.6  & 93.6  & 105.2 & 112.4  \\
    LTD-10-10 (2019) $\dag$ \cite{mao2019learning}          & 11.2                                     & 23.4                                     & 47.9                                     & 58.9                                     & 78.3                                      & 93.3                                      & 106.0                                     & 114.0                                      \\
    Hisrep (2020) $\dag$ \cite{mao2020history}              & 10.4 & 22.6 & 47.1 & 58.3 & 77.3  & 91.8  & 104.1 & 112.1  \\
    MSR-GCN (2021) $\dag$ \cite{dang2021msr}               & 11.3                                     & 24.3                                     & 50.8                                     & 61.9                                     & 80.0                                      & -                                         & -                                         & 112.9                                      \\
    ST-DGCN-10-25 (2022) $\dag$ \cite{ma2022progressively} & 10.6 & 23.1 & 47.1 & 57.9 & 76.3  & 90.7  & 102.4 & 109.7  \\
    SIMLPE (2023) \cite{guo_back_nodate} $\dag$                                                          & \multicolumn{1}{l}{ 9.6}        & \multicolumn{1}{l}{21.7}        & \multicolumn{1}{l}{46.3}                 & \multicolumn{1}{l}{57.3}                 & 75.7                & \multicolumn{1}{l}{90.1}        & \multicolumn{1}{l}{101.8}        & \multicolumn{1}{l}{109.4}        \\ 
    \midrule
    HaarMoDic(Ours)                    & \bf 9.5& \bf 21.5& \bf 45.7& \bf 56.6& \bf 75.3& \bf 89.8& \bf 101.4& \bf 109.1                                 \\
    \bottomrule
    \end{tabular}
    
    \end{table}

\subsection{Quantitative and qualitative result}
\label{quantita}
\begin{figure}[t]
    \centering
    \includegraphics[width=0.85 \textwidth]
    {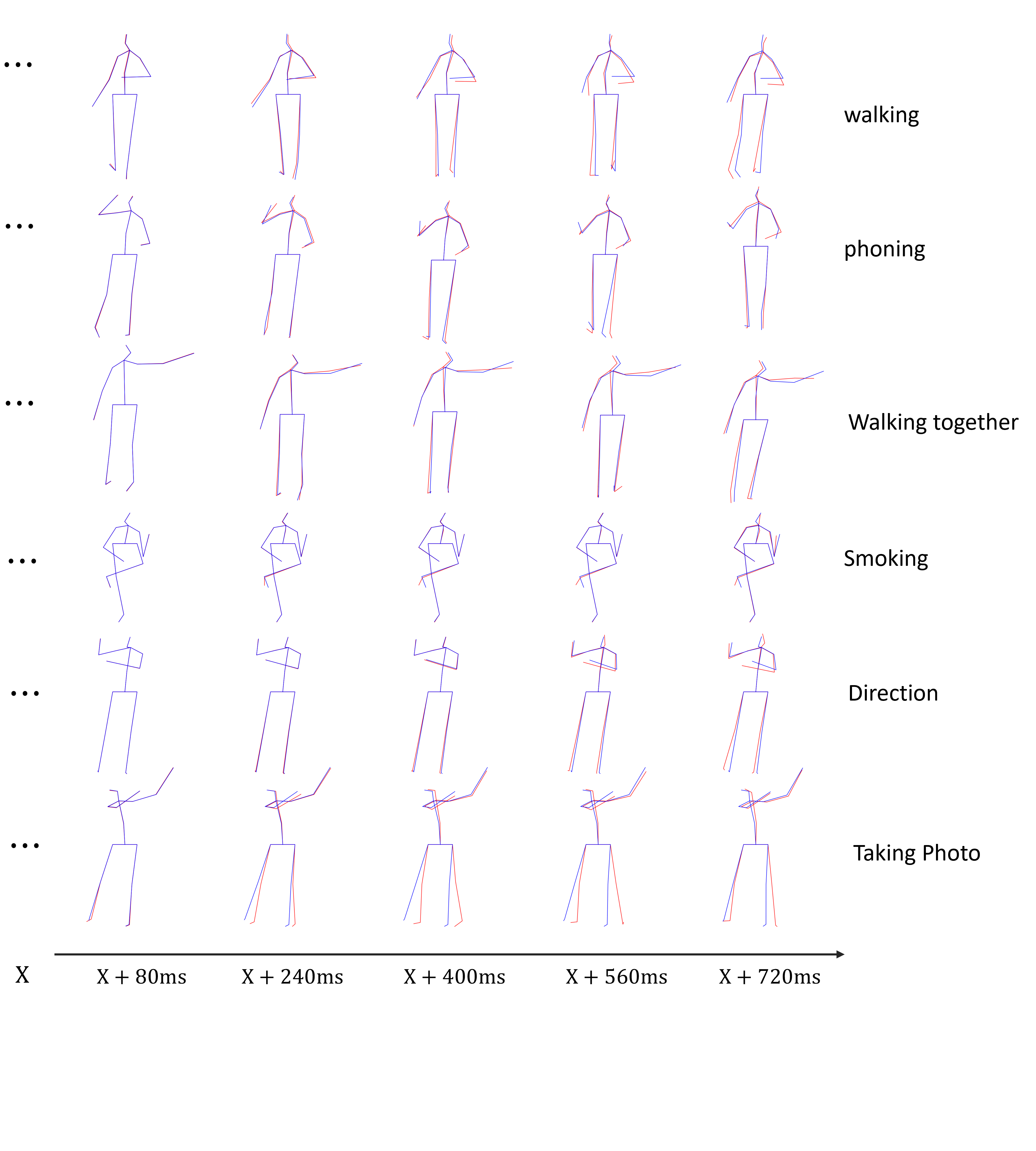}
    \caption{\textbf{Visualization of the qualitative results of our methods.} the time $X$ denotes the last frame input. $X+t$ denotes the $t$ milliseconds after the last frame where $t$ is discrete time. The skeleton in red is the grand truth, and the skeleton in blue is the result of prediction. }
    \label{fig:qualitive}
\end{figure}

\begin{table}
\tiny
\centering
\renewcommand\arraystretch{1.2}
\caption{\textbf{Action-wise results on Human3.6M} for different prediction time steps~(ms). Lower is better. 256 samples are tested for each action. $\dag$ indicates that the results are taken from the paper~\cite{mao2020history,ma2022progressively}.}
\label{table:tab_res_h36m_act_wise}
\setlength{\tabcolsep}{0.6mm}{ 
\begin{tabular}{l|cccc|cccc|cccc|cccc} 
\toprule
 Action                          & \multicolumn{4}{c|}{walking}                                                                                                                                                                                  & \multicolumn{4}{c|}{eating}                                                                                                                                                                   & \multicolumn{4}{c|}{smoking}                                                                                                                                                                          & \multicolumn{4}{c}{discussion}                                                                                                                                               \\ 
Time~(ms)                                                         & 80                                                & 400                                               & 560                                               & 1000                                              & 80                                               & 400                                       & 560                                       & 1000                                               & 80                                                 & 400                                      & 560                                               & 1000                                              & 80                                       & 400                                      & 560                                       & 1000                                       \\ 
\midrule
convSeq2Seq $\dag$ \cite{li2018convolutional}    & 17.7          & 63.6          & 72.2          & 82.3          & 11.0         & 48.4  & 61.3  & 87.1           & 11.6           & 48.9 & 60.0          & 81.7          & 17.1 & 77.6 & 98.1  & 129.3  \\
LTD-50-25 $\dag$ \cite{mao2019learning}          & 12.3                                              & 44.4                                              & 50.7                                              & 60.3                                              & 7.8                                              & 38.6                                      & 51.5                                      & 75.8                                               & 8.2                                                & 39.5                                     & 50.5                                              & 72.1                                              & 11.9                                     & 68.1                                     & 88.9                                      & 118.5                                      \\
LTD-10-10 $\dag$ \cite{mao2019learning}          & 11.1          & 42.9          & 53.1          & 70.7          & 7.0          & 37.3  & 51.1  & 78.6           & 7.5            & 37.5 & 49.4          & 71.8          & 10.8 & 65.8 & 88.1  & 121.6  \\
Hisrep $\dag$ \cite{mao2020history}              & 10.0                                              & 39.8                                              & 47.4                                              & 58.1                                              & 6.4                                              & 36.2                                      & 50.0                                      & 75.7                                               & 7.0                                                & 36.4                                     & 47.6                                              & 69.5                                              & 10.2                                     & 65.4                                     & 86.6                                      & 119.8                                      \\
MSR-GCN $\dag$ \cite{dang2021msr}               & 10.8          & 42.4          & 53.3          & 63.7          & 6.9          & 36.0  & 50.8  & 75.4           & 7.5            & 37.5 & 50.5          & 72.1          & 10.4 & 65.0 & 87.0  & 116.8  \\
ST-DGCN-10-25 $\dag$ \cite{ma2022progressively} & 11.2                                              & 42.8                                              & 49.6                                              & 58.9                                              & 6.5                                              & 36.8                                      & 50.0                                      & 74.9                                               & 7.3                                                & 37.5                                     & 48.8                                              & 69.9                                              & 10.2                                     & 64.4                                     & 86.1                                      & 116.9                                      \\
siMLPe $\dag$ \cite{ma2022progressively}        & 9.9           & \textbf{39.6} & \textbf{46.8} & \textbf{55.7} & \textbf{5.9} & 36.1  & 49.6  & 74.5           & \textbf{6.5}   & 36.3 & \textbf{47.2} & 69.3          & 9.4  & 64.3 & 85.7  & 116.3  \\ 
\midrule
HAAR(Ours)                                       & \textbf{9.8}                                      & \textbf{39.6}                                     & 47.5                                              & 56.3                                              & \textbf{5.9}                                     & \textbf{35.5}                             & \textbf{48.8}                             & \textbf{74.2}                                      & 6.6                                                & \textbf{36.0}                            & \textbf{47.2}                                     & \textbf{69.1}                                     & \textbf{9.2}                             & \textbf{62.5}                            & \textbf{83.6}                             & \textbf{114.7}                             \\ 
\toprule
 Action                          & \multicolumn{4}{c|}{directions}                                                                                                                                                                               & \multicolumn{4}{c|}{greeting}                                                                                                                                                                 & \multicolumn{4}{c|}{phoning}                                                                                                                                                                          & \multicolumn{4}{c}{posing}                                                                                                                                                   \\ 
Time~(ms)                                                         & 80                                                & 400                                               & 560                                               & 1000                                              & 80                                               & 400                                       & 560                                       & 1000                                               & 80                                                 & 400                                      & 560                                               & 1000                                              & 80                                       & 400                                      & 560                                       & 1000                                       \\ 
\midrule
convSeq2Seq $\dag$ \cite{li2018convolutional}    & 13.5          & 69.7          & 86.6          & 115.8         & 22.0         & 96.0  & 116.9 & 147.3          & 13.5           & 59.9 & 77.1          & 114.0         & 16.9 & 92.9 & 122.5 & 187.4  \\
LTD-50-25 $\dag$ \cite{mao2019learning}          & 8.8                                               & 58.0                                              & 74.2                                              & 105.5                                             & 16.2                                             & 82.6                                      & 104.8                                     & 136.8                                              & 9.8                                                & 50.8                                     & 68.8                                              & 105.1                                             & 12.2                                     & 79.9                                     & 110.2                                     & 174.8                                      \\
LTD-10-10 $\dag$ \cite{mao2019learning}          & 8.0           & 54.9          & 76.1          & 108.8         & 14.8         & 79.7  & 104.3 & 140.2          & 9.3            & 49.7 & 68.7          & 105.1         & 10.9 & 75.9 & 109.9 & 171.7  \\
Hisrep $\dag$ \cite{mao2020history}              & 7.4                                               & 56.5                                              & 73.9                                              & 106.5                                             & 13.7                                             & 78.1                                      & 101.9                                     & 138.8                                              & 8.6                                                & 49.2                                     & 67.4                                              & 105.0                                             & 10.2                                     & 75.8                                     & 107.6                                     & 178.2                                      \\
MSR-GCN $\dag$ \cite{dang2021msr}               & 7.7           & 56.2          & 75.8          & 105.9         & 15.1         & 85.4  & 106.3 & 136.3          & 9.1            & 49.8 & 67.9          & 104.7         & 10.3 & 75.9 & 112.5 & 176.5  \\
ST-DGCN-10-25 $\dag$ \cite{ma2022progressively} & 7.5                                               & 56.0                                              & 73.3                                              & \textbf{105.9 }                                   & 14.0                                             & 77.3                                      & 100.2                                     & 136.4                                              & 8.7                                                & 48.8                                     & 66.5                                              & \textbf{102.7 }                                   & 10.2                                     & 73.3                                     & 102.8                                     & 167.0                                      \\
siMLPe $\dag$ \cite{ma2022progressively}        & \textbf{6.5}  & 55.8          & \textbf{73.1} & 106.7         & 12.4         & 77.3  & 99.8  & 137.5          & \textbf{ 8.1}  & 48.6 & 66.3          & 103.3         & 8.8  & 73.8 & 103.4 & 168.7  \\ 
\midrule
HAAR(Ours)                                       & 6.6                                               & \textbf{55.2}                                     & 73.5                                              & 106.3                                             & \textbf{12.1}                                    & \textbf{75.7}                             & \textbf{97.4}                             & \textbf{133.4}                                     & \textbf{8.1}                                       & \textbf{47.6}                            & \textbf{65.6}                                     & 103.3                                             & \textbf{8.8}                             & \textbf{71.5}                            & \textbf{101.5}                            & \textbf{166.8}                             \\ 
\toprule
 Action                          & \multicolumn{4}{c|}{purchases}                                                                                                                                                                                & \multicolumn{4}{c|}{sitting}                                                                                                                                                                  & \multicolumn{4}{c|}{sittingdown}                                                                                                                                                                      & \multicolumn{4}{c}{takingphoto}                                                                                                                                              \\ 
Time~(ms)                                                         & 80                                                & 400                                               & 560                                               & 1000                                              & 80                                               & 400                                       & 560                                       & 1000                                               & 80                                                 & 400                                      & 560                                               & 1000                                              & 80                                       & 400                                      & 560                                       & 1000                                       \\ 
\midrule
convSeq2Seq $\dag$ \cite{li2018convolutional}    & 20.3          & 89.9          & 111.3         & 151.5         & 13.5         & 63.1  & 82.4  & 120.7          & 20.7           & 82.7 & 106.5         & 150.3         & 12.7 & 63.6 & 84.4  & 128.1  \\
LTD-50-25 $\dag$ \cite{mao2019learning}          & 15.2                                              & 78.1                                              & 99.2                                              & 134.9                                             & 10.4                                             & 58.3                                      & 79.2                                      & 118.7                                              & 17.1                                               & 76.4                                     & 100.2                                             & 143.8                                             & 9.6                                      & 54.3                                     & 75.3                                      & 118.8                                      \\
LTD-10-10 $\dag$ \cite{mao2019learning}          & 13.9          & 75.9          & 99.4          & 135.9         & 9.8          & 55.9  & 78.5  & 118.8          & 15.6           & 71.7 & 96.2          & 142.2         & 8.9  & 51.7 & 72.5  & 116.3  \\
Hisrep $\dag$ \cite{mao2020history}              & 13.0                                              & 73.9                                              & 95.6                                              & 134.2                                             & 9.3                                              & 56.0                                      & 76.4                                      & 115.9                                              & 14.9                                               & 72.0                                     & 97.0                                              & 143.6                                             & 8.3                                      & 51.5                                     & 72.1                                      & 115.9                                      \\
MSR-GCN $\dag$ \cite{dang2021msr}               & 13.3          & 77.8          & 99.2          & 134.5         & 9.8          & 55.5  & 77.6  & 115.9          & 15.4           & 73.8 & 102.4         & 149.4         & 8.9  & 54.4 & 77.7  & 121.9  \\
ST-DGCN-10-25 $\dag$ \cite{ma2022progressively} & 13.2                                              & 74.0                                              & 95.7                                              & \textbf{132.1}                                    & 9.1                                              & 54.6                                      & 75.1                                      & 114.8                                              & 14.7                                               & \textbf{70.0}                            & \textbf{94.4}                                     & \textbf{139.0}                                    & 8.2                                      & \textbf{50.2}                            & 70.5                                      & 112.9                                      \\
siMLPe $\dag$ \cite{ma2022progressively}        & \textbf{11.7} & 72.4          & 93.8          & 132.5         & 8.6          & 55.2  & 75.4  & 114.1          & \textbf{13.6 } & 70.8 & 95.7          & 142.4         & 7.8  & 50.8 & 71.0  & 112.8  \\ 
\midrule
HAAR(Ours)                                       & 11.9                                              & \textbf{71.9}                                     & \textbf{93.6}                                     & 133.1                                             & \textbf{8.5}                                     & \textbf{54.1}                             & \textbf{74.7}                             & \textbf{113.9}                                     & 13.7                                               & 71.0                                     & 96.6                                              & 143.7                                             & \textbf{7.7}                             & 50.3                                     & \textbf{70.2}                             & \textbf{111.9}                             \\ 
\toprule
 Action                          & \multicolumn{4}{c|}{waiting}                                                                                                                                                                                  & \multicolumn{4}{c|}{walkingdog}                                                                                                                                                               & \multicolumn{4}{c|}{walkingtogether}                                                                                                                                                                  & \multicolumn{4}{c}{average}                                                                                                                                                  \\ 
\midrule
Time~(ms)                                                         & 80                                                & 400                                               & 560                                               & 1000                                              & 80                                               & 400                                       & 560                                       & 1000                                               & 80                                                 & 400                                      & 560                                               & 1000                                              & 80                                       & 400                                      & 560                                       & 1000                                       \\ 
convSeq2Seq $\dag$ \cite{li2018convolutional}    & 14.6          & 68.7          & 87.3          & 117.7         & 27.7         & 103.3 & 122.4 & 162.4          & 15.3           & 61.2 & 72.0          & 87.4          & 16.6 & 72.7 & 90.7  & 124.2  \\
LTD-50-25 $\dag$ \cite{mao2019learning}          & 10.4                                              & 59.2                                              & 77.2                                              & 108.3                                             & 22.8                                             & 88.7                                      & 107.8                                     & 156.4                                              & 10.3                                               & 46.3                                     & 56.0                                              & 65.7                                              & 12.2                                     & 61.5                                     & 79.6                                      & 112.4                                      \\
LTD-10-10 $\dag$ \cite{mao2019learning}          & 9.2           & 54.4          & 73.4          & 107.5         & 20.9         & 86.6  & 109.7 & 150.1          & 9.6            & 44.0 & 55.7          & 69.8          & 11.2 & 58.9 & 78.3  & 114.0  \\
Hisrep $\dag$ \cite{mao2020history}              & 8.7                                               & 54.9                                              & 74.5                                              & 108.2                                             & 20.1                                             & 86.3                                      & 108.2                                     & 146.9                                              & 8.9                                                & 41.9                                     & 52.7                                              & 64.9                                              & 10.4                                     & 58.3                                     & 77.3                                      & 112.1                                      \\
MSR-GCN $\dag$ \cite{dang2021msr}               & 10.4          & 62.4          & 74.8          & 105.5         & 24.9         & 112.9 & 107.7 & 145.7          & 9.2            & 43.2 & 56.2          & 69.5          & 11.3 & 61.9 & 80.0  & 112.9  \\
ST-DGCN-10-25 $\dag$ \cite{ma2022progressively} & 8.7                                               & 53.6                                              & 71.6                                              & \textbf{103.7}                                    & 20.4                                             & 84.6                                      & 105.7                                     & 145.9                                              & 8.9                                                & 43.8                                     & 54.4                                              & 64.6                                              & 10.6                                     & 57.9                                     & 76.3                                      & 109.7                                      \\
siMLPe $\dag$ \cite{ma2022progressively}        & \textbf{7.8}  & \textbf{53.2} & \textbf{71.6} & 104.6         & 18.2         & 83.6  & 105.6 & \textbf{141.2} & \textbf{8.4}   & 41.2 & 50.8          & \textbf{61.5} & 9.6  & 57.3 & 75.7  & 109.4  \\ 
\midrule
HAAR(Ours)                                       & 7.9                                               & 54.2                                              & 73.5                                              & 105.2                                             & \textbf{17.8}                                    & \textbf{83.3}                             & \textbf{105.1}                            & 142.7                                              & 8.5                                                & \textbf{40.7}                            & \textbf{50.4}                                     & 62.1                                              & \multicolumn{1}{l}{\textbf{9.5}}         & \multicolumn{1}{l}{\textbf{56.6}}        & \multicolumn{1}{l}{\textbf{75.3}}         & \multicolumn{1}{l}{\textbf{109.1}}         \\
\bottomrule
\end{tabular}%

}

\end{table}

As shown in table \ref{table:tab_res_h36m}, the model we proposed outperforms state-of-the-art methods in every time step, specifically in the middle and middle long term. The times infer the predicted result after milliseconds after the final frames. In comparing setups, the MPJPE denotes the average Euclidean distance for all the joints. Surpassing the other methods in 1000ms and generating a precise long-term prediction is valuable for practical usage in real situations.
We visualize several prediction result at figure \ref{fig:qualitive}. Our prediction is quite close to grand truth even in long term.

Several previous methods make a trade-off between short-term and long-term prediction, which appears that later methods surpass the previous works on one side but fail on the other. Our method exceeds the previous works in every testing interval because of the benefit of Haar transform rather than trade-off performance between short- and long-term prediction.  

As for the detailed results for each type of human motion, they are visualized at table \ref{table:tab_res_h36m_act_wise}. It is clear that we surpass the other method in most of the items with most of the time steps, which contributes to our average results. Furthermore, the proposed method distinguishly elaborates on the quality of the complex cases, including posing, greeting, and walking dogs. This is the benefit of our Multi-resolution architecture, which capture hints from motion sequence from higher resolution than other methods. 

\subsection{Ablation study}
\label{ablation}
  \textbf{Ablation on each main module}. \quad We ablate implementation of each main module in HaarMoDic in Table \ref{tab:main_modules}. Repeating the last frame of the given sequence is set as the baseline, which is picked from \cite{guo_back_nodate}. We first add a pre-processing module only to observe its function. Its function turns out to be limited. Then, we add the DCT transform module over that, and the result remains almost unchanged. Finally, we added our  Multi-Resolution Haar Blocks into the network, and the result became acceptable. It shows that our MR-Haar block brings most of the effectiveness into this method.

\begin{table}[]
\renewcommand\arraystretch{1.2}
\centering
\caption{\textbf{Ablation on main modules.} \dag denotes the result for repeating the last frame of the given sequence, which is taken from \cite{guo_back_nodate}. We ablate the DCT module (DCT), Multi-Resolution Haar module (MR), and pre-process and post-process module (Pre/Pose). We use the MPJPE as our metric and calculate it at 8 different time intervals.}

\begin{tabular}{ccc|cccccccc}
\toprule
\multicolumn{3}{c|}{Modules}                  & \multicolumn{8}{c}{Time Seteps (ms)}                      \\ 
DCT           & MR            & Pre/Post      & 80   & 160  & 320  & 400  & 560   & 720   & 880   & 1000  \\  \midrule
\multicolumn{3}{c|}{Repeating Last-Frame\dag} & 23.8 & 44.4 & 76.1 & 88.2 & 107.4 & 121.6 & 131.6 & 136.6 \\
              &               & \checkmark    & 23.8 & 44.1 & 75.6 & 87.8 & 106.5 & 120.7 & 130.7 & 135.7 \\
\checkmark    &               & \checkmark    & 23.8 & 44.1 & 75.6 & 87.7 & 106.5 & 120.6 & 130.6 & 135.6 \\

\checkmark    & \checkmark    & \checkmark    & 9.5  & 21.5 & 45.7 & 56.6 & 75.3  & 89.8  & 101.4 & 109.1 \\ \bottomrule
\end{tabular}
\label{tab:main_modules}
\end{table}

\textbf{Data augmentation.} \quad In Table.\ref{ablation:aug}, we evaluate of the contribution of data augmentation. \textbf{flip} denotes randomly flipping the Y-axis of a motion sequence. \textbf{reverse} denotes for the reverse the sequence in the temporal axis, i.e., rewind the sequence. The random flip and reverse in time sequence improve the results. Meanwhile, flipping the coordinate of a sequence is slightly less effective than reversing the sequence. An interesting observation is that the final improvement is the addition of the separate improvement of each augmentation, which is important for our method.

\begin{table}[]
\renewcommand\arraystretch{1.2}
\centering
    \caption{\textbf{Ablation of Data augmentation.} The result of ablate different types of data augmentation.\textbf{flip} means to flip the Y-axis of the motion with a possibility. In contrast, \textbf{Reverse} means reverse the time sequence of the snippet of the motion with a possibility.}
    \begin{tabular}{cc|cccccccc}
    \toprule
    \multicolumn{2}{c|}{Aug} & \multicolumn{8}{c}{Time(millisecond)}                           \\ 
    Flip        & Reverse    & 80  & 160  & 320  & 400  & 560  & 720  & 880   & 1000  \\ \midrule
                &            & 9.9 & 22.1 & 47.2 & 58.3 & 77.0 & 91.4 & 103.5 & 111.3 \\
    \checkmark  &            & 9.7 & 21.9 & 46.7 & 57.8 & 76.5 & 90.8 & 102.6 & 110.2 \\
                & \checkmark & 9.6 & 21.6 & 45.9 & 56.9 & 75.6 & 90.1 & 102.2 & 110.0 \\

    \checkmark  & \checkmark & 9.5 & 21.5 & 45.7 & 56.6 & 75.3 & 89.8 & 101.4 & 109.1 \\ \bottomrule
    \end{tabular}

\label{ablation:aug}
\end{table}

\textbf{Evaluation on the inner parameter of the MR-Haar block} \quad In Table \ref{tab:inner}, we analyze the contribution of parameter setting inside the Multi-resolution Haar block to the final performance. The \textbf{length} denotes the number of FC blocks in a going-through scenario, while the block denotes how many MR-Haar blocks we implement. It is clear that if we keep the length as 3, and increase the block number, we could get better results than baseline. However, if the depth of the block is too deep, it becomes hard to train the network, which drawbacks the performance. It is similar to the length parameter; when it grows from zero, it increases the performance. However, if the length is too long, we cannot reach the best performance. We choose to implement 4 blocks, each using 3 FC-block to conduct a going-through scenario.
\begin{table}[]
\centering
\renewcommand\arraystretch{1.2}
\caption{\textbf{Ablation of the architecture.} \textbf{block} denotes the number of the MR-Haar blocks we implement in the network, and \textbf{length} denotes the number of full-connected layers in the MLP sequence.
}
\begin{tabular}{cc|cccccccc}
\toprule
block & length & \multicolumn{8}{c}{MPJPE at time(millisecond)}                     \\ \midrule
0     & 0      & 23.8 & 44.1 & 75.6 & 87.7 & 106.5 & 120.6 & 130.6 & 135.6 \\
1     & 3     & 9.9  & 22.3 & 47.1 & 58.1 & 77.5  & 92.9  & 104.7 & 112.2 \\
2     & 3     & 9.8  & 22.1 & 46.7 & 57.7 & 76.9  & 92.1  & 104.5 & 112.3 \\
8     & 3     & 10.3 & 22.8 & 47.3 & 58.2 & 76.8  & 91.2  & 103.1 & 111.0 \\
4     & 1      & 9.8  & 22.3 & 46.9 & 57.8 & 76.9  & 92.3  & 104.5 & 112.1 \\
4     & 2      & 9.6  & 21.8 & 46.3 & 57.3 & 76.2  & 91.1  & 103.1 & 110.9 \\
4     & 6     & 9.6  & 21.5 & 45.8 & 56.7 & 75.3  & 90.5  & 102.8 & 110.4 \\

4     & 3     & 9.5  & 21.5 & 45.7 & 56.6 & 75.3  & 89.8  & 101.4 & 109.1 \\ \bottomrule
\end{tabular}

\label{tab:inner}
\end{table}

\begin{table}
\renewcommand\arraystretch{1.2}
    \centering
    \caption{\textbf{Ablation of DCT and layer normalization} We ablate the Layer Normalization (LN) inside the FC block and the DCT module. We use the MPJPE as the metric and calculate it at 8 different time steps.}
    \begin{tabular}{c|cccccccc}
    \toprule
    Ablation  & 80  & 160  & 320  & 400  & 560  & 720  & 880   & 1000                    \\ \midrule
    w/o LN    & 12.9 & 29.3 & 62.5 & 76.3 & 97.4 & 111.6 & 121.5 & 127.2 \\
    w/o DCT   & 9.9  & 22.1 & 46.5 & 57.5 & 76.2 & 90.6  & 102.4 & 110.2 \\ \midrule

    HaarMoDic(ours) & 9.5  & 21.5 & 45.7 & 56.6 & 75.3 & 89.8  & 101.4 & 109.1 \\ \bottomrule
    \end{tabular}
    
    \label{ablation:LN_DCT}
    \end{table}

\textbf{DCT and Layer Normalization} \quad In Table.\ref{ablation:LN_DCT}, we analyze the contribution from discrete cosine transforms and layer normalization. It shows that adopting DCT and layer normalization can improve the result, especially for the Layer Normalization (LN) operation inside the FC blocks, which is vital for the final performance. The DCT module does increase the performance, but it is less vital than the Layer Normalization operation. 

\section{Conclusion}
This paper proposes HaarMoDic, an enhanced spatial-temporal multi-resolution network for human motion prediction. The major component is the MR-Haar block. By utilizing the Haar transform, the network can access information in spatial-temporal dimensions simultaneously which previous works undergo. The reported ablation study also illustrates the contribution of the module above. Our approach outperforms state-of-the-art methods on the Human3.6M benchmark dataset. We hope to further explore the multi-resolution approaches and architecture that can better release the power of attention mechanisms in the future.









\bibliography{sn-article}


\begin{thebibliography}{39}
\ifx \bisbn   \undefined \def \bisbn  #1{ISBN #1}\fi
\ifx \binits  \undefined \def \binits#1{#1}\fi
\ifx \bauthor  \undefined \def \bauthor#1{#1}\fi
\ifx \batitle  \undefined \def \batitle#1{#1}\fi
\ifx \bjtitle  \undefined \def \bjtitle#1{#1}\fi
\ifx \bvolume  \undefined \def \bvolume#1{\textbf{#1}}\fi
\ifx \byear  \undefined \def \byear#1{#1}\fi
\ifx \bissue  \undefined \def \bissue#1{#1}\fi
\ifx \bfpage  \undefined \def \bfpage#1{#1}\fi
\ifx \blpage  \undefined \def \blpage #1{#1}\fi
\ifx \burl  \undefined \def \burl#1{\textsf{#1}}\fi
\ifx \doiurl  \undefined \def \doiurl#1{\url{https://doi.org/#1}}\fi
\ifx \betal  \undefined \def \betal{\textit{et al.}}\fi
\ifx \binstitute  \undefined \def \binstitute#1{#1}\fi
\ifx \binstitutionaled  \undefined \def \binstitutionaled#1{#1}\fi
\ifx \bctitle  \undefined \def \bctitle#1{#1}\fi
\ifx \beditor  \undefined \def \beditor#1{#1}\fi
\ifx \bpublisher  \undefined \def \bpublisher#1{#1}\fi
\ifx \bbtitle  \undefined \def \bbtitle#1{#1}\fi
\ifx \bedition  \undefined \def \bedition#1{#1}\fi
\ifx \bseriesno  \undefined \def \bseriesno#1{#1}\fi
\ifx \blocation  \undefined \def \blocation#1{#1}\fi
\ifx \bsertitle  \undefined \def \bsertitle#1{#1}\fi
\ifx \bsnm \undefined \def \bsnm#1{#1}\fi
\ifx \bsuffix \undefined \def \bsuffix#1{#1}\fi
\ifx \bparticle \undefined \def \bparticle#1{#1}\fi
\ifx \barticle \undefined \def \barticle#1{#1}\fi
\bibcommenthead
\ifx \bconfdate \undefined \def \bconfdate #1{#1}\fi
\ifx \botherref \undefined \def \botherref #1{#1}\fi
\ifx \url \undefined \def \url#1{\textsf{#1}}\fi
\ifx \bchapter \undefined \def \bchapter#1{#1}\fi
\ifx \bbook \undefined \def \bbook#1{#1}\fi
\ifx \bcomment \undefined \def \bcomment#1{#1}\fi
\ifx \oauthor \undefined \def \oauthor#1{#1}\fi
\ifx \citeauthoryear \undefined \def \citeauthoryear#1{#1}\fi
\ifx \endbibitem  \undefined \def \endbibitem {}\fi
\ifx \bconflocation  \undefined \def \bconflocation#1{#1}\fi
\ifx \arxivurl  \undefined \def \arxivurl#1{\textsf{#1}}\fi
\csname PreBibitemsHook\endcsname

\bibitem[\protect\citeauthoryear{Makris et~al.}{2016}]{makris2016augmented}
\begin{barticle}
\bauthor{\bsnm{Makris}, \binits{S.}},
\bauthor{\bsnm{Karagiannis}, \binits{P.}},
\bauthor{\bsnm{Koukas}, \binits{S.}},
\bauthor{\bsnm{Matthaiakis}, \binits{A.-S.}}:
\batitle{Augmented reality system for operator support in human--robot collaborative assembly}.
\bjtitle{CIRP Annals}
\bvolume{65}(\bissue{1}),
\bfpage{61}--\blpage{64}
(\byear{2016})
\end{barticle}
\endbibitem

\bibitem[\protect\citeauthoryear{Honda et~al.}{2020}]{honda2020rnn}
\begin{bchapter}
\bauthor{\bsnm{Honda}, \binits{Y.}},
\bauthor{\bsnm{Kawakami}, \binits{R.}},
\bauthor{\bsnm{Naemura}, \binits{T.}}:
\bctitle{Rnn-based motion prediction in competitive fencing considering interaction between players.}
In: \bbtitle{BMVC}
(\byear{2020})
\end{bchapter}
\endbibitem

\bibitem[\protect\citeauthoryear{Gulzar et~al.}{2021}]{Gulzar2021Survey}
\begin{barticle}
\bauthor{\bsnm{Gulzar}, \binits{M.}},
\bauthor{\bsnm{Muhammad}, \binits{Y.}},
\bauthor{\bsnm{Muhammad}, \binits{N.}}:
\batitle{A survey on motion prediction of pedestrians and vehicles for autonomous driving}.
\bjtitle{IEEE Access}
\bvolume{9},
\bfpage{137957}--\blpage{137969}
(\byear{2021})
\doiurl{10.1109/ACCESS.2021.3118224}
\end{barticle}
\endbibitem

\bibitem[\protect\citeauthoryear{Sun et~al.}{2022}]{sun2022p4p}
\begin{botherref}
\oauthor{\bsnm{Sun}, \binits{Q.}},
\oauthor{\bsnm{Huang}, \binits{X.}},
\oauthor{\bsnm{Williams}, \binits{B.C.}},
\oauthor{\bsnm{Zhao}, \binits{H.}}:
P4P: Conflict-Aware Motion Prediction for Planning in Autonomous Driving
(2022)
\end{botherref}
\endbibitem

\bibitem[\protect\citeauthoryear{Martinez et~al.}{2017}]{martinez_human_2017}
\begin{bchapter}
\bauthor{\bsnm{Martinez}, \binits{J.}},
\bauthor{\bsnm{Black}, \binits{M.J.}},
\bauthor{\bsnm{Romero}, \binits{J.}}:
\bctitle{On human motion prediction using recurrent neural networks}.
In: \bbtitle{2017 {IEEE} Conference on Computer Vision and Pattern Recognition ({CVPR})},
pp. \bfpage{4674}--\blpage{4683}
(\byear{2017}).
\doiurl{10.1109/CVPR.2017.497} .
\burl{http://ieeexplore.ieee.org/document/8099980/}
Accessed 2023-08-12
\end{bchapter}
\endbibitem

\bibitem[\protect\citeauthoryear{Fragkiadaki et~al.}{2015}]{fragkiadaki_recurrent_2015}
\begin{bchapter}
\bauthor{\bsnm{Fragkiadaki}, \binits{K.}},
\bauthor{\bsnm{Levine}, \binits{S.}},
\bauthor{\bsnm{Felsen}, \binits{P.}},
\bauthor{\bsnm{Malik}, \binits{J.}}:
\bctitle{Recurrent network models for human dynamics}.
In: \bbtitle{2015 {IEEE} International Conference on Computer Vision ({ICCV})},
pp. \bfpage{4346}--\blpage{4354}
(\byear{2015}).
\doiurl{10.1109/ICCV.2015.494} .
\burl{http://ieeexplore.ieee.org/document/7410851/}
\end{bchapter}
\endbibitem

\bibitem[\protect\citeauthoryear{Cui et~al.}{2020}]{cui2020learning}
\begin{bchapter}
\bauthor{\bsnm{Cui}, \binits{Q.}},
\bauthor{\bsnm{Sun}, \binits{H.}},
\bauthor{\bsnm{Yang}, \binits{F.}}:
\bctitle{{Learning Dynamic Relationships for 3D Human Motion Prediction}}.
In: \bbtitle{Conference on Computer Vision and Pattern Recognition}
(\byear{2020})
\end{bchapter}
\endbibitem

\bibitem[\protect\citeauthoryear{Hochreiter and Schmidhuber}{1997}]{hochreiter1997long}
\begin{barticle}
\bauthor{\bsnm{Hochreiter}, \binits{S.}},
\bauthor{\bsnm{Schmidhuber}, \binits{J.}}:
\batitle{Long short-term memory}.
\bjtitle{Neural computation}
\bvolume{9}(\bissue{8}),
\bfpage{1735}--\blpage{1780}
(\byear{1997})
\end{barticle}
\endbibitem

\bibitem[\protect\citeauthoryear{Chung et~al.}{2014}]{chung2014empirical}
\begin{botherref}
\oauthor{\bsnm{Chung}, \binits{J.}},
\oauthor{\bsnm{Gulcehre}, \binits{C.}},
\oauthor{\bsnm{Cho}, \binits{K.}},
\oauthor{\bsnm{Bengio}, \binits{Y.}}:
Empirical evaluation of gated recurrent neural networks on sequence modeling.
arXiv preprint arXiv:1412.3555
(2014)
\end{botherref}
\endbibitem

\bibitem[\protect\citeauthoryear{Guo et~al.}{2023}]{guo_back_nodate}
\begin{botherref}
\oauthor{\bsnm{Guo}, \binits{W.}},
\oauthor{\bsnm{Du}, \binits{Y.}},
\oauthor{\bsnm{Shen}, \binits{X.}},
\oauthor{\bsnm{Lepetit}, \binits{V.}},
\oauthor{\bsnm{Alameda-Pineda}, \binits{X.}},
\oauthor{\bsnm{Moreno-Noguer}, \binits{F.}}:
Back to {MLP}: A simple baseline for human motion prediction.
WACV2023
(2023)
\end{botherref}
\endbibitem

\bibitem[\protect\citeauthoryear{Mao et~al.}{2019}]{mao_learning_2019}
\begin{bchapter}
\bauthor{\bsnm{Mao}, \binits{W.}},
\bauthor{\bsnm{Liu}, \binits{M.}},
\bauthor{\bsnm{Salzmann}, \binits{M.}},
\bauthor{\bsnm{Li}, \binits{H.}}:
\bctitle{Learning trajectory dependencies for human motion prediction}.
In: \bbtitle{2019 {IEEE}/{CVF} International Conference on Computer Vision ({ICCV})},
pp. \bfpage{9488}--\blpage{9496}
(\byear{2019}).
\doiurl{10.1109/ICCV.2019.00958} .
\burl{https://ieeexplore.ieee.org/document/9009559/}
Accessed 2023-08-12
\end{bchapter}
\endbibitem

\bibitem[\protect\citeauthoryear{Dang et~al.}{2021}]{dang_msr-gcn_2021}
\begin{bchapter}
\bauthor{\bsnm{Dang}, \binits{L.}},
\bauthor{\bsnm{Nie}, \binits{Y.}},
\bauthor{\bsnm{Long}, \binits{C.}},
\bauthor{\bsnm{Zhang}, \binits{Q.}},
\bauthor{\bsnm{Li}, \binits{G.}}:
\bctitle{{MSR}-{GCN}: Multi-scale residual graph convolution networks for human motion prediction}.
In: \bbtitle{2019 {IEEE}/{CVF} International Conference on Computer Vision ({ICCV})},
pp. \bfpage{11447}--\blpage{11456}
(\byear{2021}).
\doiurl{10.1109/ICCV48922.2021.01127}
\end{bchapter}
\endbibitem

\bibitem[\protect\citeauthoryear{Mao et~al.}{2020}]{mao2020history}
\begin{bchapter}
\bauthor{\bsnm{Mao}, \binits{W.}},
\bauthor{\bsnm{Liu}, \binits{M.}},
\bauthor{\bsnm{Salzmann}, \binits{M.}}:
\bctitle{{History Repeats Itself: Human Motion Prediction via Motion Attention}}.
In: \bbtitle{European Conference on Computer Vision}
(\byear{2020})
\end{bchapter}
\endbibitem

\bibitem[\protect\citeauthoryear{Aksan et~al.}{2021}]{aksan2021spatio}
\begin{bchapter}
\bauthor{\bsnm{Aksan}, \binits{E.}},
\bauthor{\bsnm{Kaufmann}, \binits{M.}},
\bauthor{\bsnm{Cao}, \binits{P.}},
\bauthor{\bsnm{Hilliges}, \binits{O.}}:
\bctitle{A spatio-temporal transformer for 3d human motion prediction}.
In: \bbtitle{2021 International Conference on 3D Vision (3DV)},
pp. \bfpage{565}--\blpage{574}
(\byear{2021}).
\bcomment{IEEE}
\end{bchapter}
\endbibitem

\bibitem[\protect\citeauthoryear{Ionescu et~al.}{2013}]{ionescu2013human3}
\begin{botherref}
\oauthor{\bsnm{Ionescu}, \binits{C.}},
\oauthor{\bsnm{Papava}, \binits{D.}},
\oauthor{\bsnm{Olaru}, \binits{V.}},
\oauthor{\bsnm{Sminchisescu}, \binits{C.}}:
{Human3. 6m: Large Scale Datasets and Predictive Methods for 3D Human Sensing in Natural Environments}.
IEEE Transactions on Pattern Analysis and Machine Intelligence
(2013)
\end{botherref}
\endbibitem

\bibitem[\protect\citeauthoryear{Fragkiadaki et~al.}{2015}]{fragkiadaki2015recurrent}
\begin{bchapter}
\bauthor{\bsnm{Fragkiadaki}, \binits{K.}},
\bauthor{\bsnm{Levine}, \binits{S.}},
\bauthor{\bsnm{Felsen}, \binits{P.}},
\bauthor{\bsnm{Malik}, \binits{J.}}:
\bctitle{{Recurrent Network Models for Human Dynamics}}.
In: \bbtitle{International Conference on Computer Vision}
(\byear{2015})
\end{bchapter}
\endbibitem

\bibitem[\protect\citeauthoryear{Jain et~al.}{2016}]{jain2016structural}
\begin{bchapter}
\bauthor{\bsnm{Jain}, \binits{A.}},
\bauthor{\bsnm{Zamir}, \binits{A.R.}},
\bauthor{\bsnm{Savarese}, \binits{S.}},
\bauthor{\bsnm{Saxena}, \binits{A.}}:
\bctitle{{Structural-Rnn: Deep Learning on Spatio-Temporal Graphs}}.
In: \bbtitle{Conference on Computer Vision and Pattern Recognition}
(\byear{2016})
\end{bchapter}
\endbibitem

\bibitem[\protect\citeauthoryear{Liu et~al.}{2019}]{liu2019towards}
\begin{bchapter}
\bauthor{\bsnm{Liu}, \binits{Z.}},
\bauthor{\bsnm{Wu}, \binits{S.}},
\bauthor{\bsnm{Jin}, \binits{S.}},
\bauthor{\bsnm{Liu}, \binits{Q.}},
\bauthor{\bsnm{Lu}, \binits{S.}},
\bauthor{\bsnm{Zimmermann}, \binits{R.}},
\bauthor{\bsnm{Cheng}, \binits{L.}}:
\bctitle{{Towards Natural and Accurate Future Motion Prediction of Humans and Animals}}.
In: \bbtitle{Conference on Computer Vision and Pattern Recognition}
(\byear{2019})
\end{bchapter}
\endbibitem

\bibitem[\protect\citeauthoryear{Martinez et~al.}{2017}]{martinez2017human}
\begin{bchapter}
\bauthor{\bsnm{Martinez}, \binits{J.}},
\bauthor{\bsnm{Black}, \binits{M.J.}},
\bauthor{\bsnm{Romero}, \binits{J.}}:
\bctitle{{On Human Motion Prediction Using Recurrent Neural Networks}}.
In: \bbtitle{Conference on Computer Vision and Pattern Recognition}
(\byear{2017})
\end{bchapter}
\endbibitem

\bibitem[\protect\citeauthoryear{Chiu et~al.}{2019}]{chiu_action-agnostic_2019}
\begin{bchapter}
\bauthor{\bsnm{Chiu}, \binits{H.-K.}},
\bauthor{\bsnm{Adeli}, \binits{E.}},
\bauthor{\bsnm{Wang}, \binits{B.}},
\bauthor{\bsnm{Huang}, \binits{D.-A.}},
\bauthor{\bsnm{Niebles}, \binits{J.C.}}:
\bctitle{Action-agnostic human pose forecasting}.
In: \bbtitle{2019 {IEEE} Winter Conference on Applications of Computer Vision ({WACV})},
pp. \bfpage{1423}--\blpage{1432}
(\byear{2019}).
\doiurl{10.1109/WACV.2019.00156}
\end{bchapter}
\endbibitem

\bibitem[\protect\citeauthoryear{Guo et~al.}{2022}]{guomulti}
\begin{bchapter}
\bauthor{\bsnm{Guo}, \binits{W.}},
\bauthor{\bsnm{Bie}, \binits{X.}},
\bauthor{\bsnm{Alameda-Pineda}, \binits{X.}},
\bauthor{\bsnm{Moreno-Noguer}, \binits{F.}}:
\bctitle{{Multi-Person Extreme Motion Prediction}}.
In: \bbtitle{Conference on Computer Vision and Pattern Recognition}
(\byear{2022})
\end{bchapter}
\endbibitem

\bibitem[\protect\citeauthoryear{Li et~al.}{2020}]{li2020dynamic}
\begin{bchapter}
\bauthor{\bsnm{Li}, \binits{M.}},
\bauthor{\bsnm{Chen}, \binits{S.}},
\bauthor{\bsnm{Zhao}, \binits{Y.}},
\bauthor{\bsnm{Zhang}, \binits{Y.}},
\bauthor{\bsnm{Wang}, \binits{Y.}},
\bauthor{\bsnm{Tian}, \binits{Q.}}:
\bctitle{{Dynamic Multiscale Graph Neural Networks for 3D Skeleton Based Human Motion Prediction}}.
In: \bbtitle{Conference on Computer Vision and Pattern Recognition}
(\byear{2020})
\end{bchapter}
\endbibitem

\bibitem[\protect\citeauthoryear{Mao et~al.}{2019}]{mao2019learning}
\begin{bchapter}
\bauthor{\bsnm{Mao}, \binits{W.}},
\bauthor{\bsnm{Liu}, \binits{M.}},
\bauthor{\bsnm{Salzmann}, \binits{M.}},
\bauthor{\bsnm{Li}, \binits{H.}}:
\bctitle{{Learning Trajectory Dependencies for Human Motion Prediction}}.
In: \bbtitle{International Conference on Computer Vision}
(\byear{2019})
\end{bchapter}
\endbibitem

\bibitem[\protect\citeauthoryear{Ma et~al.}{2022}]{ma2022progressively}
\begin{bchapter}
\bauthor{\bsnm{Ma}, \binits{T.}},
\bauthor{\bsnm{Nie}, \binits{Y.}},
\bauthor{\bsnm{Long}, \binits{C.}},
\bauthor{\bsnm{Zhang}, \binits{Q.}},
\bauthor{\bsnm{Li}, \binits{G.}}:
\bctitle{{Progressively Generating Better Initial Guesses Towards Next Stages for High-Quality Human Motion Prediction}}.
In: \bbtitle{Conference on Computer Vision and Pattern Recognition}
(\byear{2022})
\end{bchapter}
\endbibitem

\bibitem[\protect\citeauthoryear{Li et~al.}{2021}]{li2021symbiotic}
\begin{botherref}
\oauthor{\bsnm{Li}, \binits{M.}},
\oauthor{\bsnm{Chen}, \binits{S.}},
\oauthor{\bsnm{Chen}, \binits{X.}},
\oauthor{\bsnm{Zhang}, \binits{Y.}},
\oauthor{\bsnm{Wang}, \binits{Y.}},
\oauthor{\bsnm{Tian}, \binits{Q.}}:
{Symbiotic Graph Neural Networks for 3D Skeleton-Based Human Action Recognition and Motion Prediction}.
IEEE Transactions on Pattern Analysis and Machine Intelligence
(2021)
\end{botherref}
\endbibitem

\bibitem[\protect\citeauthoryear{Lehrmann et~al.}{2014}]{lehrmann2014efficient}
\begin{bchapter}
\bauthor{\bsnm{Lehrmann}, \binits{A.M.}},
\bauthor{\bsnm{Gehler}, \binits{P.V.}},
\bauthor{\bsnm{Nowozin}, \binits{S.}}:
\bctitle{{Efficient Nonlinear Markov Models for Human Motion}}.
In: \bbtitle{Conference on Computer Vision and Pattern Recognition}
(\byear{2014})
\end{bchapter}
\endbibitem

\bibitem[\protect\citeauthoryear{Wang et~al.}{2005}]{wang2005gaussian}
\begin{bchapter}
\bauthor{\bsnm{Wang}, \binits{J.M.}},
\bauthor{\bsnm{Fleet}, \binits{D.J.}},
\bauthor{\bsnm{Hertzmann}, \binits{A.}}:
\bctitle{{Gaussian Process Dynamical Models}}.
In: \bbtitle{Advances in Neural Information Processing Systems}
(\byear{2005})
\end{bchapter}
\endbibitem

\bibitem[\protect\citeauthoryear{Taylor et~al.}{2007}]{taylor2007modeling}
\begin{bchapter}
\bauthor{\bsnm{Taylor}, \binits{G.W.}},
\bauthor{\bsnm{Hinton}, \binits{G.E.}},
\bauthor{\bsnm{Roweis}, \binits{S.T.}}:
\bctitle{{Modeling Human Motion Using Binary Latent Variables}}.
In: \bbtitle{Advances in Neural Information Processing Systems}
(\byear{2007})
\end{bchapter}
\endbibitem

\bibitem[\protect\citeauthoryear{Cai et~al.}{2020}]{cai2020learning}
\begin{bchapter}
\bauthor{\bsnm{Cai}, \binits{Y.}},
\bauthor{\bsnm{Huang}, \binits{L.}},
\bauthor{\bsnm{Wang}, \binits{Y.}},
\bauthor{\bsnm{Cham}, \binits{T.-J.}},
\bauthor{\bsnm{Cai}, \binits{J.}},
\bauthor{\bsnm{Yuan}, \binits{J.}},
\bauthor{\bsnm{Liu}, \binits{J.}},
\bauthor{\bsnm{Yang}, \binits{X.}},
\bauthor{\bsnm{Zhu}, \binits{Y.}},
\bauthor{\bsnm{Shen}, \binits{X.}},
\bauthor{\bsnm{Others}}:
\bctitle{{Learning Progressive Joint Propagation for Human Motion Prediction}}.
In: \bbtitle{European Conference on Computer Vision}
(\byear{2020})
\end{bchapter}
\endbibitem

\bibitem[\protect\citeauthoryear{Butepage et~al.}{2017}]{butepage2017deep}
\begin{bchapter}
\bauthor{\bsnm{Butepage}, \binits{J.}},
\bauthor{\bsnm{Black}, \binits{M.J.}},
\bauthor{\bsnm{Kragic}, \binits{D.}},
\bauthor{\bsnm{Kjellstrom}, \binits{H.}}:
\bctitle{{Deep Representation Learning for Human Motion Prediction and Classification}}.
In: \bbtitle{Conference on Computer Vision and Pattern Recognition}
(\byear{2017})
\end{bchapter}
\endbibitem

\bibitem[\protect\citeauthoryear{B\"utepage et~al.}{2018}]{butepage2018anticipating}
\begin{bchapter}
\bauthor{\bsnm{B\"utepage}, \binits{J.}},
\bauthor{\bsnm{Kjellstr\"om}, \binits{H.}},
\bauthor{\bsnm{Kragic}, \binits{D.}}:
\bctitle{{Anticipating Many Futures: Online Human Motion Prediction and Generation for Human-Robot Interaction}}.
In: \bbtitle{International Conference on Robotics and Automation}
(\byear{2018})
\end{bchapter}
\endbibitem

\bibitem[\protect\citeauthoryear{Hernandez et~al.}{2019}]{hernandeziccv2019}
\begin{bchapter}
\bauthor{\bsnm{Hernandez}, \binits{A.}},
\bauthor{\bsnm{Gall}, \binits{J.}},
\bauthor{\bsnm{Moreno-Noguer}, \binits{F.}}:
\bctitle{{Human Motion Prediction via Spatio-Temporal Inpainting}}.
In: \bbtitle{International Conference on Computer Vision}
(\byear{2019})
\end{bchapter}
\endbibitem

\bibitem[\protect\citeauthoryear{Li et~al.}{2018}]{li2018convolutional}
\begin{bchapter}
\bauthor{\bsnm{Li}, \binits{C.}},
\bauthor{\bsnm{Zhang}, \binits{Z.}},
\bauthor{\bsnm{Lee}, \binits{W.S.}},
\bauthor{\bsnm{Lee}, \binits{G.H.}}:
\bctitle{{Convolutional Sequence to Sequence Model for Human Dynamics}}.
In: \bbtitle{Conference on Computer Vision and Pattern Recognition}
(\byear{2018})
\end{bchapter}
\endbibitem

\bibitem[\protect\citeauthoryear{Gui et~al.}{2018}]{gui2018adversarial}
\begin{bchapter}
\bauthor{\bsnm{Gui}, \binits{L.-Y.}},
\bauthor{\bsnm{Wang}, \binits{Y.-X.}},
\bauthor{\bsnm{Liang}, \binits{X.}},
\bauthor{\bsnm{Moura}, \binits{J.M.}}:
\bctitle{{Adversarial Geometry-Aware Human Motion Prediction}}.
In: \bbtitle{European Conference on Computer Vision}
(\byear{2018})
\end{bchapter}
\endbibitem

\bibitem[\protect\citeauthoryear{Mao et~al.}{}]{mao_history_2020}
\begin{botherref}
\oauthor{\bsnm{Mao}, \binits{W.}},
\oauthor{\bsnm{Liu}, \binits{M.}},
\oauthor{\bsnm{Salzmann}, \binits{M.}}:
History repeats itself: Human motion prediction via motion attention.
In: Computer Vision–{ECCV} 2020:,
pp. 474--489.
Springer
\end{botherref}
\endbibitem

\bibitem[\protect\citeauthoryear{Ma et~al.}{2022}]{ma_progressively_2022}
\begin{bchapter}
\bauthor{\bsnm{Ma}, \binits{T.}},
\bauthor{\bsnm{Nie}, \binits{Y.}},
\bauthor{\bsnm{Long}, \binits{C.}},
\bauthor{\bsnm{Zhang}, \binits{Q.}},
\bauthor{\bsnm{Li}, \binits{G.}}:
\bctitle{Progressively generating better initial guesses towards next stages for high-quality human motion prediction}.
In: \bbtitle{2022 {IEEE}/{CVF} Conference on Computer Vision and Pattern Recognition ({CVPR})},
pp. \bfpage{6427}--\blpage{6436}
(\byear{2022}).
\doiurl{10.1109/CVPR52688.2022.00633} .
\burl{https://ieeexplore.ieee.org/document/9879362/}
Accessed 2023-07-30
\end{bchapter}
\endbibitem

\bibitem[\protect\citeauthoryear{Cotter}{2020}]{cotter_2020}
\begin{botherref}
\oauthor{\bsnm{Cotter}, \binits{F.}}:
Uses of complex wavelets in deep convolutional neural networks
(2020)
\doiurl{10.17863/CAM.53748}
\end{botherref}
\endbibitem

\bibitem[\protect\citeauthoryear{Wang et~al.}{2020}]{wang2020deep}
\begin{barticle}
\bauthor{\bsnm{Wang}, \binits{J.}},
\bauthor{\bsnm{Sun}, \binits{K.}},
\bauthor{\bsnm{Cheng}, \binits{T.}},
\bauthor{\bsnm{Jiang}, \binits{B.}},
\bauthor{\bsnm{Deng}, \binits{C.}},
\bauthor{\bsnm{Zhao}, \binits{Y.}},
\bauthor{\bsnm{Liu}, \binits{D.}},
\bauthor{\bsnm{Mu}, \binits{Y.}},
\bauthor{\bsnm{Tan}, \binits{M.}},
\bauthor{\bsnm{Wang}, \binits{X.}}, \betal:
\batitle{Deep high-resolution representation learning for visual recognition}.
\bjtitle{IEEE transactions on pattern analysis and machine intelligence}
\bvolume{43}(\bissue{10}),
\bfpage{3349}--\blpage{3364}
(\byear{2020})
\end{barticle}
\endbibitem

\bibitem[\protect\citeauthoryear{Dang et~al.}{2021}]{dang2021msr}
\begin{bchapter}
\bauthor{\bsnm{Dang}, \binits{L.}},
\bauthor{\bsnm{Nie}, \binits{Y.}},
\bauthor{\bsnm{Long}, \binits{C.}},
\bauthor{\bsnm{Zhang}, \binits{Q.}},
\bauthor{\bsnm{Li}, \binits{G.}}:
\bctitle{{MSR-GCN: Multi-Scale Residual Graph Convolution Networks for Human Motion Prediction}}.
In: \bbtitle{International Conference on Computer Vision}
(\byear{2021})
\end{bchapter}
\endbibitem

\end{thebibliography}
\end{document}